\documentclass{svproc}
\usepackage{graphicx}
\usepackage{subcaption}
\usepackage{float}
\usepackage{amsmath}
\usepackage{booktabs}
\usepackage{tabularx}
\usepackage{url}

\graphicspath{{figures/}}
\begin{document}
\mainmatter              
\title{Unsupervised Roofline Extraction from True Orthophotos for LoD2 Building Model Reconstruction}
\titlerunning{Unsupervised Roofline Extraction from True Orthophotos}  
%
\author{Weixiao Gao\inst{1}, Ravi Peters\inst{2}, \and Jantien Stoter\inst{1}}
\authorrunning{Weixiao Gao et al.} 
%
\tocauthor{Weixiao Gao, Ravi Peters, Jantien Stoter}
\institute{Dept. Urbanism, Delft University of Technology, The Netherlands,\\
\email{(w.gao-1, j.e.stoter)@tudelft.nl},\\
\and
3DGI, Zoetermeer, The Netherlands,\\
\email{ravi.peters@3dgi.nl}}


\maketitle              

\begin{abstract}
This paper discusses the reconstruction of LoD2 building models from 2D and 3D data for large-scale urban environments. Traditional methods involve the use of LiDAR point clouds, but due to high costs and long intervals associated with acquiring such data for rapidly developing areas, researchers have started exploring the use of point clouds generated from (oblique) aerial images. However, using such point clouds for traditional plane detection-based methods can result in significant errors and introduce noise into the reconstructed building models. To address this, this paper presents a method for extracting rooflines from true orthophotos using line detection for the reconstruction of building models at the LoD2 level. The approach is able to extract relatively complete rooflines without the need for pre-labeled training data or pre-trained models. These lines can directly be used in the LoD2 building model reconstruction process. The method is superior to existing plane detection-based methods and state-of-the-art deep learning methods in terms of the accuracy and completeness of the reconstructed building. Our source code is available at \url{https://github.com/tudelft3d/Roofline-extraction-from-orthophotos}.

\end{abstract}

\section{Introduction}\label{introduction}
With the recent advancements in computer vision and photogrammetry technology, the acquisition of both 2D and 3D data for large-scale urban environments has become feasible. 
This has led to the derivation of urban 2D data from street view, aerial, and satellite images, while 3D data is derived from LiDAR point clouds, dense image matching generated point clouds, and textured meshes. Building models in 3D urban scenes are essential for various applications such as solar irradiation~\cite{biljecki2015propagation,besuievsky2018skyline}, photovoltaic analysis~\cite{catita2014extending,eerenstein2015tessera}, building illumination analysis~\cite{saran2015citygml}, and urban planning~\cite{Chen2011,czynska2014application}. 
However, for large-scale urban scene analysis, it is often necessary to find the optimal balance between simplicity and preserving essential, detailed geometric features. 
To this end, LoD2 building models, conforming to the CityGML standard~\cite{kolbe2021citygml}, have gained popularity.

The automatic reconstruction of LoD2 building models from 2D and 3D data has been a topic of active research. 
Traditionally, point clouds obtained from airborne LiDAR have been used to generate such models~\cite{peters2022automated}. 
However, due to the high costs and long intervals associated with acquiring and collecting LiDAR point clouds for large-scale urban scenes, it can be challenging to capture rapidly developing and changing areas in cities. 
Consequently, there may be missing data for these regions, which makes it difficult to update the models accurately. 
To address this issue, researchers have started exploring the use of point clouds generated from aerial images with the dense image matching pipelines for automatic LoD2 building model reconstruction~\cite{wang2023reconstruction}. 
Dense image matching point clouds offer several advantages over LiDAR point clouds, including being easily obtainable, cost-effective, and having color information. 
Furthermore, they can be frequently updated to reflect changes in the urban environment, making them more suitable for vector building reconstruction tasks in large-scale, high-frequency urban scenes where the data must be updated frequently to reflect urban development. 
As a result, dense image matching point clouds have become an area of interest for researchers studying automated LoD2 building model reconstruction.
\begin{figure}[H]
	\centering
	\begin{subfigure}[t]{0.23\textwidth}
		\includegraphics[width=\linewidth]{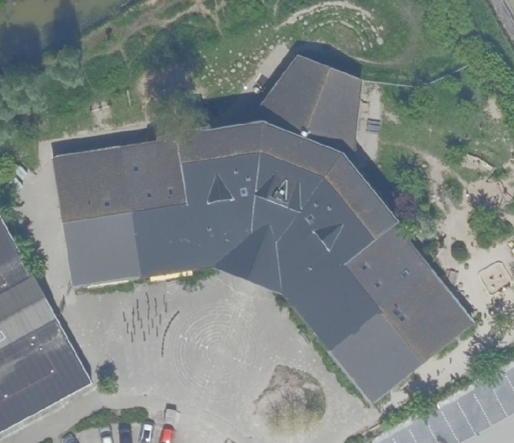}
		\caption{}
	\end{subfigure}
	\hspace*{\fill}
	\begin{subfigure}[t]{0.23\textwidth}
		\includegraphics[width=\linewidth]{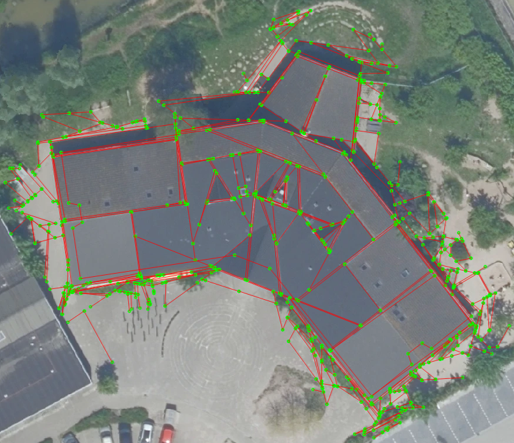}
		\caption{}
	\end{subfigure}
	\hspace*{\fill}
	\begin{subfigure}[t]{0.23\textwidth}
		\includegraphics[width=\linewidth]{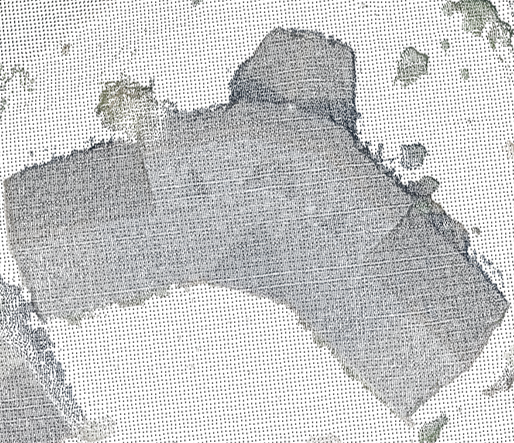}
		\caption{}
	\end{subfigure}
	\hspace*{\fill}
	\begin{subfigure}[t]{0.23\textwidth}
		\includegraphics[width=\linewidth]{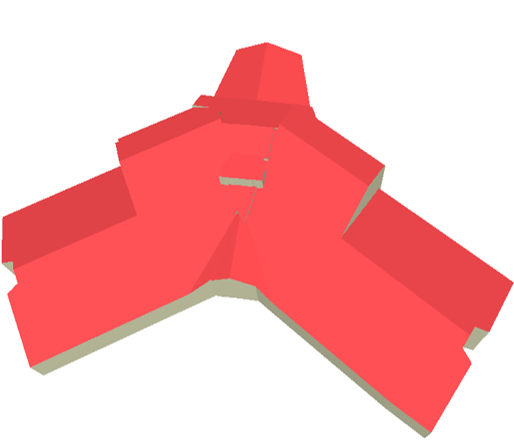}
		\caption{}
	\end{subfigure}
	\caption{The workflow of our method. 
		 First, we use a true orthophoto (a) as input. Next, we perform line extraction (b) to partition the building footprint, which generates separate roof parts. We then utilize a dense point cloud (c) to extrude the partition results and reconstruct a LoD2 building model (d).} 
	\label{fig:pipeline}
\end{figure}
Generating dense point clouds from aerial oblique photogrammetry images can result in significant errors and introduce noise into the generated point clouds, compared to high-precision LiDAR point clouds. 
This can make originally planar regions, such as roof planes of buildings, become uneven, which creates challenges for reconstructing LoD2 building models. 
Traditional point cloud plane detection-based methods~\cite{schnabel2007efficient,lafarge2012creating} can also exacerbate the issue caused by dense image matching point clouds by increasing complexity and reducing accuracy in the reconstructed models. 
However, the original images used to generate these point clouds and the synthesized orthoimagery typically contain less noise and fewer errors, providing an opportunity to extract roof structures directly from the images and combine them with height information from the point clouds extracted from the same source images for reconstructing high-precision LoD2 building models.

In recent years, more and more research has focused on how to use deep learning methods to extract roof structures from images~\cite{alidoost20192d,nauata2020vectorizing,zhang2020conv,zhang2021structured,zhao2022extracting,chen2022heat}. Although these methods have performed well on public datasets~\cite{nauata2020vectorizing}, they still require a large amount of manually labeled training data and cannot guarantee good generalization ability. More importantly, they cannot ensure the geometric completeness of the extracted roof structures, especially for complex buildings.

This paper presents a method for extracting rooflines using line detection, which can be used for the reconstruction of building models at the Lod2 level Figure~\ref{fig:pipeline}. As our results will show, our approach has the advantage of being able to extract relatively complete roof lines without the need for any pre-labeled training data or pre-trained models. Moreover, our method surpasses both traditional plane detection-based methods and state-of-the-art deep learning methods based on transformers in terms of both the accuracy and completeness of the reconstructed building. Our method is applicable to large-scale urban scenes and can be used for extracting roof structures of complex buildings.

\section{Recent Advances in Roofline Extraction}\label{related_work}

In recent years, deep learning methods have received increasing attention for extracting roof structures from images using neural networks. For instance, Fatemeh Alidoost et al. (2019)~\cite{alidoost20192d} proposed a CNN-based approach for 3D building reconstruction from a single 2D image, which achieved accurate height prediction, roofline segmentation, and building boundary extraction. However, the accuracy of the method degraded for test data with different spatial-spectral characteristics and complicated buildings. Another pioneering work by Nelson Nauata et al. (2020)~\cite{nauata2020vectorizing} used CNNs to detect geometric primitives and infer their relationships, fusing all information into a planar graph through holistic geometric reasoning for reconstructing a building architecture. This work made significant improvements over the existing state-of-the-art but has limitations in handling missed corners, curved buildings, and weak image signals.

Furthermore, Fuyang Zhang et al. (2020)~\cite{zhang2020conv} proposed Conv-MPN, a message passing neural architecture, for structured outdoor architecture reconstruction, which achieved significant performance improvements over existing prior-free solutions but has the drawback of extensive memory consumption and limitations on the number of corner candidates. They further improved the method by presenting a novel explore-and-classify framework~\cite{zhang2021structured} for structured outdoor architecture reconstruction. The method learns to classify the correctness of primitives while exploring the space of reconstructions via heuristic actions, and demonstrated significant improvements over all state-of-the-art methods, with a few limitations related to slow test-time inference and corner detection failures in extreme cases. Another similar work is presented by Wufan Zhao et al. (2022)~\cite{zhao2022extracting}, who proposed RSGNN, an end-to-end learning framework for planar roof structure extraction in polygon format from VHR remote sensing images, which combines primitive detectors and GNN-based relationship inference and shows superior results in both qualitative and quantitative evaluations. However, the method has limitations in handling complex roof structures and labeling accuracy of reference data.

A recent work HEAT is proposed by Jiacheng Chen et al. (2022)~\cite{chen2022heat}, which is an attention-based neural network for reconstructing a planar graph from 2D raster images. It uses transformer decoders and iterative inference at test time but still faces challenges in missing corners and rare structures. Despite their advantages, deep learning methods require a large amount of training data with manual labels and may not ensure good generalization and geometric completeness for extracting roof structures, especially for complex buildings. Even a single missing roof line can result in significant geometric errors in the reconstructed building models.

Johann Lussange et al. (2023)~\cite{lussange2022sateroof} recently proposed a novel approach for extracting roof sections from satellite images to reconstruct LoD2 building models. Their method employs deep learning with a Mask R-CNN model to segment roof sections in 2D satellite images and uses panoptic segmentation to infer heights-to-ground for full 3D reconstruction. While this method shows potential for large-scale LoD2 building reconstruction, it has limitations, as it heavily relies on hand-annotated training data and struggles with complex roof structures.

\section{Methodology}\label{method}
Our method involves three main steps: 1) building image cropping; 2) extracting line segments from the building images; and 3) reconstructing LoD2 building models.

\subsection{Building image cropping}\label{preprocess}
Due to the large size of the original true orthophoto and the presence of other objects in addition to buildings, direct line detection would extract many redundant line segments. Therefore, our first objective is to extract a single building image for each building from the true orthophoto. 
\begin{figure}[H]
	\centering
	\begin{subfigure}[t]{0.33\textwidth}
		\includegraphics[width=\linewidth]{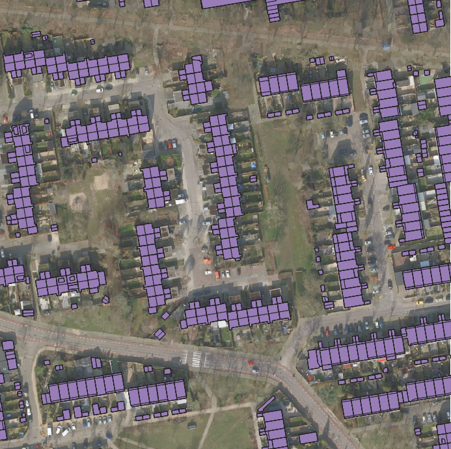}
		\caption{}
		\label{fig:building_crpping_a}
	\end{subfigure}
	\hspace*{\fill}
	\begin{subfigure}[t]{0.33\textwidth}
		\includegraphics[width=\linewidth]{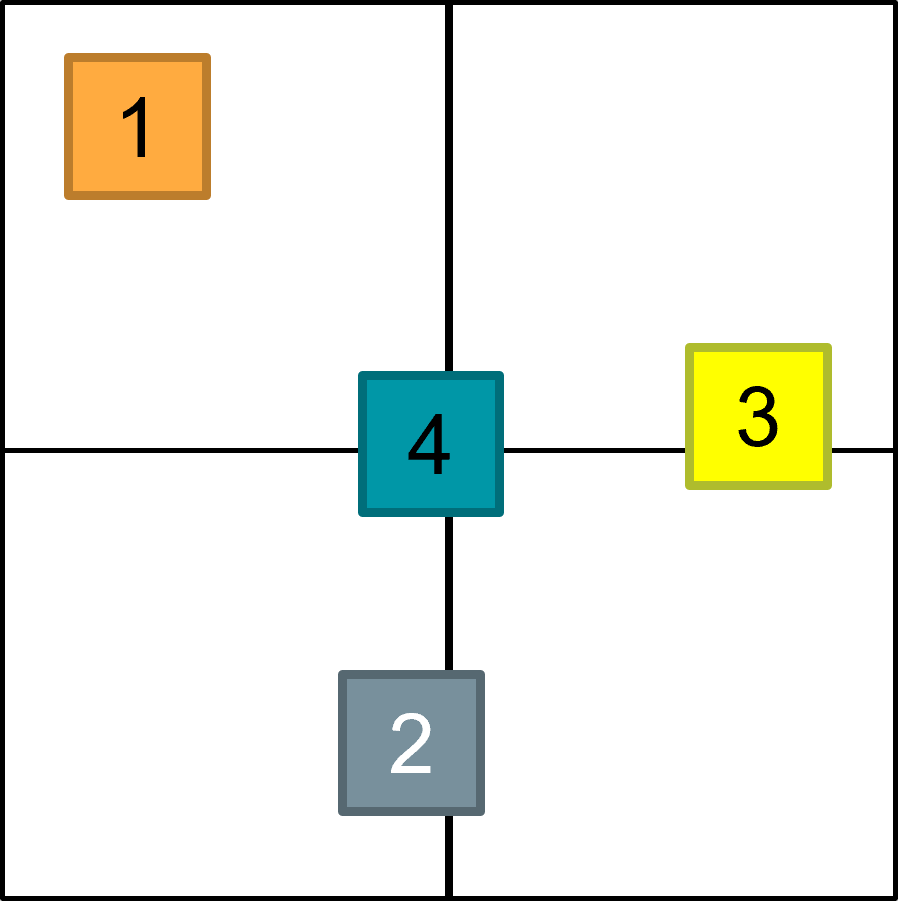}
		\caption{}
		\label{fig:building_crpping_b}
	\end{subfigure}
	\hspace*{\fill}
	\begin{subfigure}[t]{0.28\textwidth}
		\includegraphics[width=\linewidth]{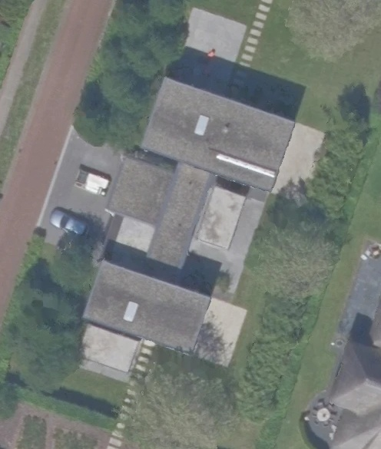}
		\caption{}
		\label{fig:building_crpping_c}
	\end{subfigure}
	\caption{The workflow of building image cropping. 
		(a) Building footprints on the map.
		(b) Four types of correspondences that can exist between a bounding box of a building and true orthophotos.
		(c) Cropped building image.} 
	\label{fig:building_crpping}
\end{figure}

We use GIS data, specifically the building footprints on the map (see Figure~\ref{fig:building_crpping_a}), to assist with the extraction process. First, we merge adjacent buildings according to their footprints. Next, we establish a bounding box based on the footprint of a single building and determine the corresponding region in the true orthophotos by using the coordinates of the four corners of the bounding box (as illustrated in Figure~\ref{fig:building_crpping_b}). The geographic coordinates of the top-left corner of the true orthophoto are already known. The correspondences between a specific bounding box and orthophotos can be divided into four cases: 1) a single image, 2) two images joined horizontally, 3) two images joined vertically, and 4) four images joined both horizontally and vertically. In the final step, we use the building boundary boxes to crop and stitch the corresponding images, generating a single composite image for each building (see Figure~\ref{fig:building_crpping_b}). These images will serve as input data for the next stage of line segment detection.

\subsection{Line Extraction}\label{roofline_extraction}
\begin{figure}[H]
	\centering
	\begin{subfigure}[t]{0.25\textwidth}
		\includegraphics[width=\linewidth]{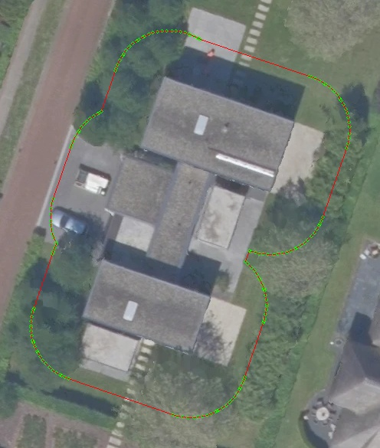}
		\caption{}
		\label{fig:line_extraction_a}
	\end{subfigure}
	\hspace*{\fill}
	\begin{subfigure}[t]{0.25\textwidth}
		\includegraphics[width=\linewidth]{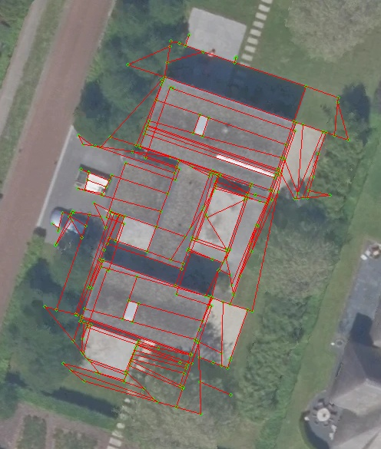}
		\caption{}
		\label{fig:line_extraction_b}
	\end{subfigure}
	\hspace*{\fill}
	\begin{subfigure}[t]{0.4\textwidth}
		\includegraphics[width=\linewidth]{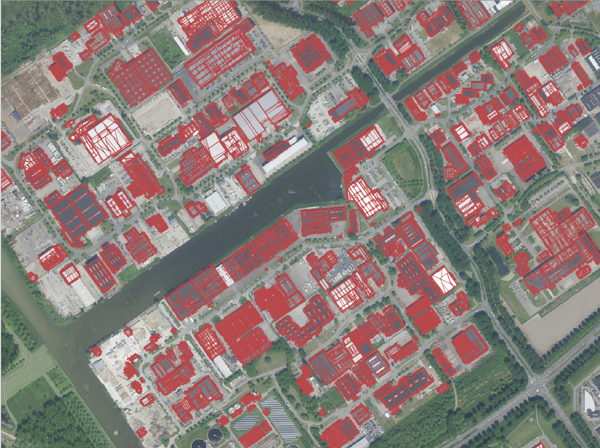}
		\caption{}
		\label{fig:line_extraction_c}
	\end{subfigure}
	\caption{Line extraction workflow on the single building image. 
		(a) Buffered building footprints.
		(b) Cropped line segments.
		(c) Georeferenced rooflines.} 
	\label{fig:line_extraction}
\end{figure}

Our objective is to detect line segments in the single building image obtained in the previous step and convert the detection results from image coordinates to geographic coordinates. We use the KIPPI~\cite{bauchet2018kippi} algorithm as our primary tool for line detection. Firstly, it employs a line segment detector to identify line segments in the image. These detected line segments are then globally regularized based on geometric properties like parallelism, perpendicularity, and collinearity. Finally, the line segments are bidirectionally extended using kinetic data structures to segment the image. As shown in Figure~\ref{fig:line_extraction_b}, rooflines represent a subset of these segmented lines. 

However, the single building image may include redundant information such as non-building features surrounding the building. To filter out such information, we use a buffer (of 60 pixels in all experiments) generated from the building footprint to refine the segmented lines (see Figure~\ref{fig:line_extraction_a}). Next, we convert the coordinates of each node of every line segment in the image to geographic coordinates using Equation~\ref{eq:equa1} as follows:
\begin{equation}
\begin{aligned}
g_x = t_{x} + (p_{x} + b_{x}) \times s\\
g_y = t_{y} - (p_{y} + b_{y}) \times s
\end{aligned}\label{eq:equa1}
\end{equation}
where $g_x$ and $g_y$ represent the geographic coordinates of the line node, while $p_x$ and $p_y$ represent its pixel coordinates. The $t_x$ and $t_y$ denote the geographic coordinates of the top left corner of the corresponding true orthophoto. The $s$ represents the pixel size. The results provide geometric information for the next step of building reconstruction (see~Figure~\ref{fig:line_extraction_c}).

It is important to highlight that despite the removal of redundant lines, certain artifacts may remain in the roof structure. Nonetheless, in the context of 3D building model reconstruction, prioritizing roofline completeness takes precedence over eliminating redundancy. The presence of real roof lines is crucial for accurate reconstruction, as the absence of such lines can lead to inaccuracies in the final model. Any remaining redundant lines can be effectively addressed in the subsequent reconstruction step through cell selection with the aid of an elevation prior~\cite{peters2022automated}.

\subsection{LoD2 building Reconstruction}\label{reconstruction}
In this step, our goal is to reconstruct LoD2 building models using the set of line segments extracted in the previous step, the building footprint, and the point cloud generated by the dense image matching pipeline from the same source images. Our approach is based on the method proposed by Ravi Peters et al. (2022)~\cite{peters2022automated}, with the main difference being the replacement of the line segments generated through point cloud detection of planes with the set of line segments extracted from the image. We then use the set of line segments extracted from the image to subdivide the building footprint and obtain the roof plane structure. Finally, based on the height information provided by the point cloud, we extrude the roof parts to generate a watertight and 2D-manifold solid building model in LoD2. Since we rely on the building footprint to extract the building point clouds and filter rooflines, it is possible that the reconstructed models may not capture roof overhangs accurately, potentially leading to their omission.

\section{Experimental Results}\label{exp_results}
\subsection{Dataset}\label{dataset}
We utilized an orthophoto dataset with a resolution of 8cm~\cite{luchtorthofoto}, covering three cities including Almere, Den Haag, and Leiden. The building footprint data was obtained from the Building and Address Register of the Netherlands (BAG). The dense point cloud was generated using oblique aerial images with Nframes SURE software by a workflow designed and executed by the Dutch Kadaster. Our experimental setup consisted of a computer equipped with an AMD Ryzen Threadripper 1920X 12-Core Processor and 32GB of memory.

\subsection{Evaluation and Comparison}\label{eval}
Our method for line extraction was tested on three cities in the Netherlands, with extraction times of 1.62 hours for 27,395 buildings in Almere, 3.22 hours for 43,520 buildings in Den Haag, and 2.85 hours for 53,502 buildings in Leiden. The time required for line extraction in these cities is mainly influenced by the number of buildings and the complexity of their external geometric structures.

\begin{figure}[!tb]
	\centering
	\begin{subfigure}[t]{0.32\textwidth}
		\includegraphics[width=\linewidth]{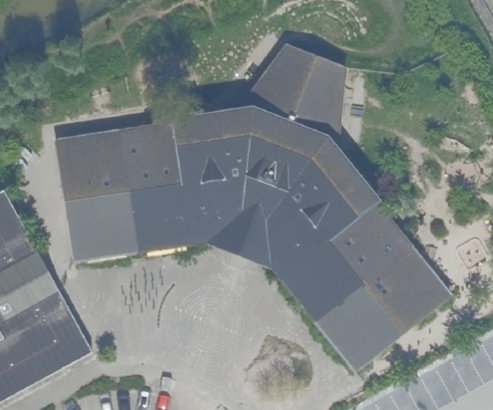}
		\label{fig:roofline_cmp_1a}
	\end{subfigure}
	\hspace*{\fill}
	\begin{subfigure}[t]{0.32\textwidth}
		\includegraphics[width=\linewidth]{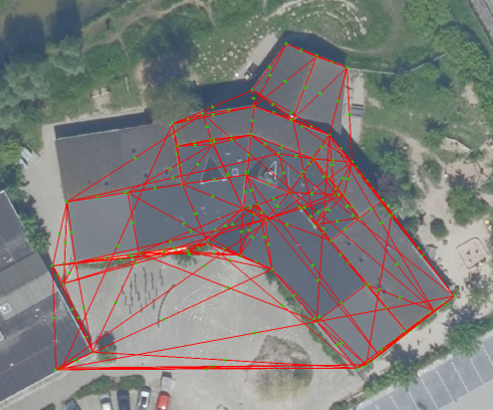}
		\label{fig:roofline_cmp_1b}
	\end{subfigure}
	\hspace*{\fill}
	\begin{subfigure}[t]{0.32\textwidth}
		\includegraphics[width=\linewidth]{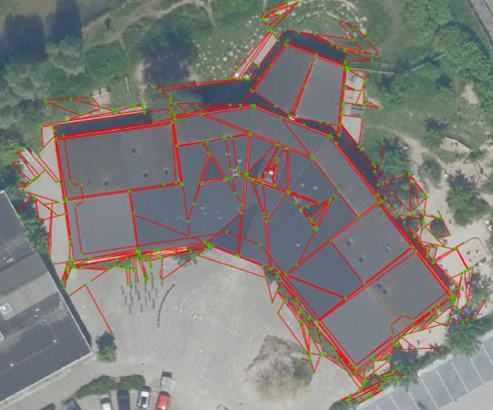}
		\label{fig:roofline_cmp_1c}
	\end{subfigure}

	\begin{subfigure}[t]{0.32\textwidth}
		\includegraphics[width=\linewidth]{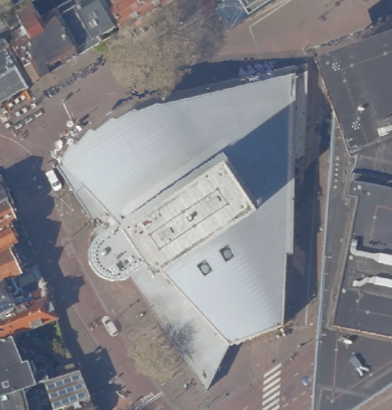}
		\label{fig:roofline_cmp_2a}
	\end{subfigure}
	\hspace*{\fill}
	\begin{subfigure}[t]{0.32\textwidth}
		\includegraphics[width=\linewidth]{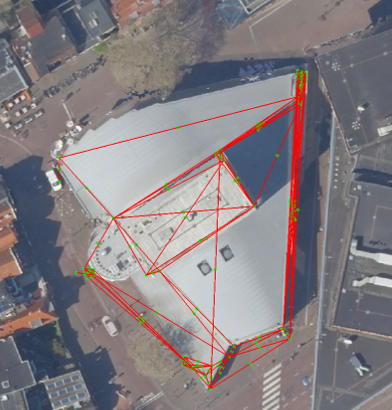}
		\label{fig:roofline_cmp_2b}
	\end{subfigure}
	\hspace*{\fill}
	\begin{subfigure}[t]{0.32\textwidth}
		\includegraphics[width=\linewidth]{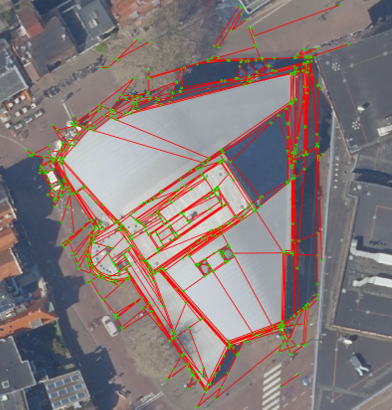}
		\label{fig:roofline_cmp_2c}
	\end{subfigure}

	\begin{subfigure}[t]{0.32\textwidth}
		\includegraphics[width=\linewidth]{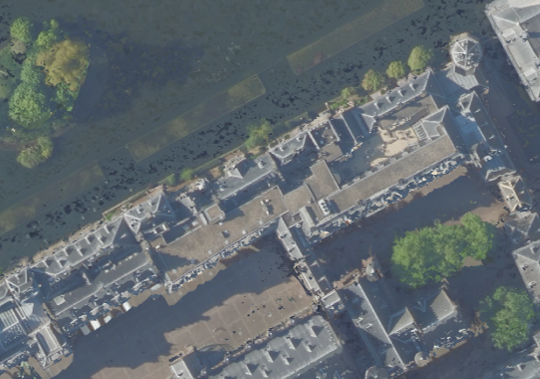}
		\caption{}
		\label{fig:roofline_cmp_3a}
	\end{subfigure}
	\hspace*{\fill}
	\begin{subfigure}[t]{0.32\textwidth}
		\includegraphics[width=\linewidth]{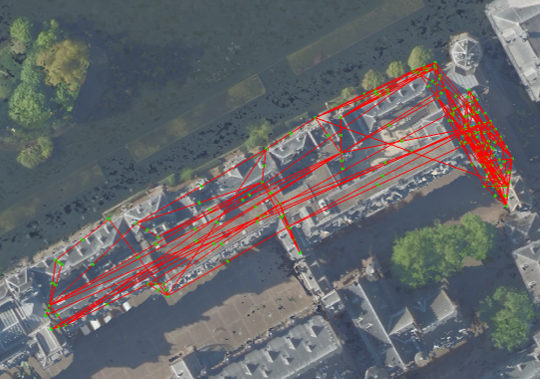}
		\caption{}
		\label{fig:roofline_cmp_3b}
	\end{subfigure}
	\hspace*{\fill}
	\begin{subfigure}[t]{0.32\textwidth}
		\includegraphics[width=\linewidth]{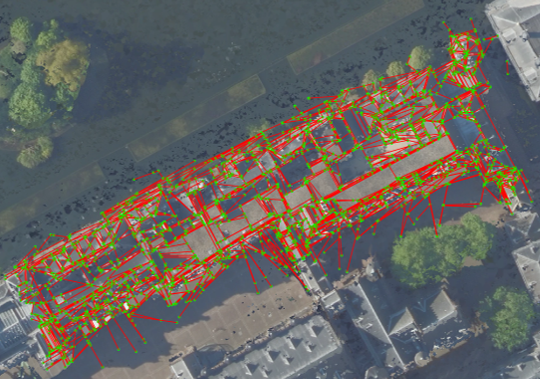}
		\caption{}
		\label{fig:roofline_cmp_3c}
	\end{subfigure}
	\caption{Roofline comparison. 
		(a) Input building image.
		(b) HEAT results.
		(c) KIPPI results.} 
	\label{fig:roofline_cmp}
\end{figure}

\paragraph{\textbf{Evaluation of rooflines.}}\label{roofline_eval}
We selected twelve representative buildings from the three cities and we projected the rooflines extracted from the 3D BAG building models~\cite{peters2022automated} onto the orthophotos as a reference.
We compared our roof line detection method (i.e., KIPPI method) with the HEAT method~\cite{chen2022heat}, which currently holds the top position in public benchmark datasets. To predict the roof structure of the generated single-building images (as shown in Figure 4), we used the trained model provided by the authors of HEAT~\cite{chen2022heat}. 
We calculated the completeness of our method by measuring the percentage of extracted lines that overlapped with the reference lines (within 25 pixels offset in all experiments). Our method achieved around $90\%$ overlap with the reference lines, while the HEAT method only achieved about $44\%$.
This comparison clearly shows that our method produces more complete results, which is further evident in the subsequent 3D reconstruction outcomes. In addition, deep learning-based methods require fixed-size image inputs due to limited GPU memory, and HEAT~\cite{chen2022heat} limits the size of each building image to $256 \times 256$. This limitation results in a significant loss of image resolution when scaling building images, especially for individual large and complex buildings (see Figure~\ref{fig:roofline_cmp}).

\begin{figure}[!tb]
	\centering
	\begin{subfigure}[t]{0.18\textwidth}
		\includegraphics[width=\linewidth]{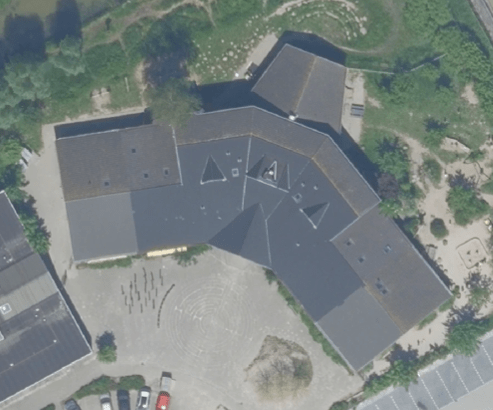}
		\label{fig:lod2_cmp_1a}
	\end{subfigure}
	\hspace*{\fill}
	\begin{subfigure}[t]{0.18\textwidth}
		\includegraphics[width=\linewidth]{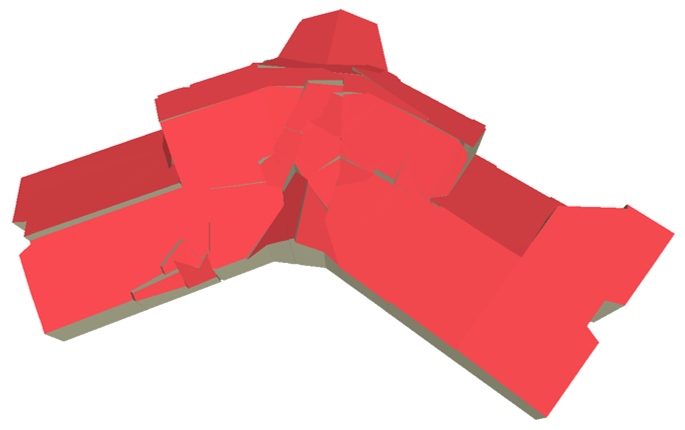}
		\label{fig:lod2_cmp_1b}
	\end{subfigure}
	\hspace*{\fill}
	\begin{subfigure}[t]{0.18\textwidth}
		\includegraphics[width=\linewidth]{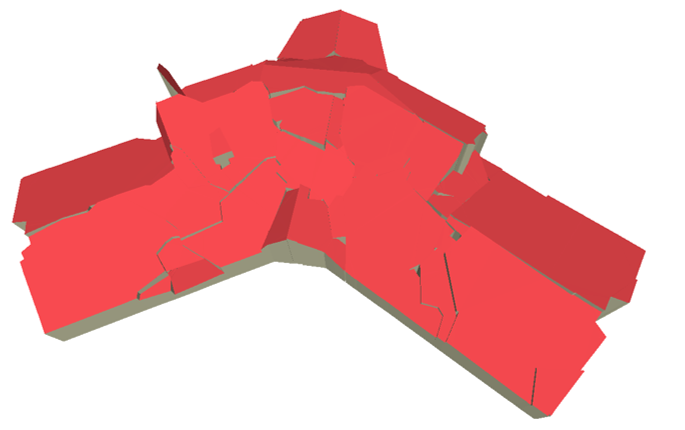}
		\label{fig:lod2_cmp_1c}
	\end{subfigure}
	\hspace*{\fill}
	\begin{subfigure}[t]{0.18\textwidth}
		\includegraphics[width=\linewidth]{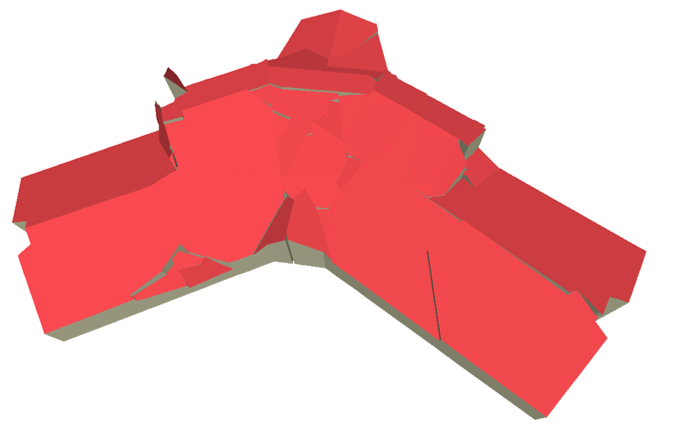}
		\label{fig:lod2_cmp_1d}
	\end{subfigure}
	\hspace*{\fill}
	\begin{subfigure}[t]{0.18\textwidth}
		\includegraphics[width=\linewidth]{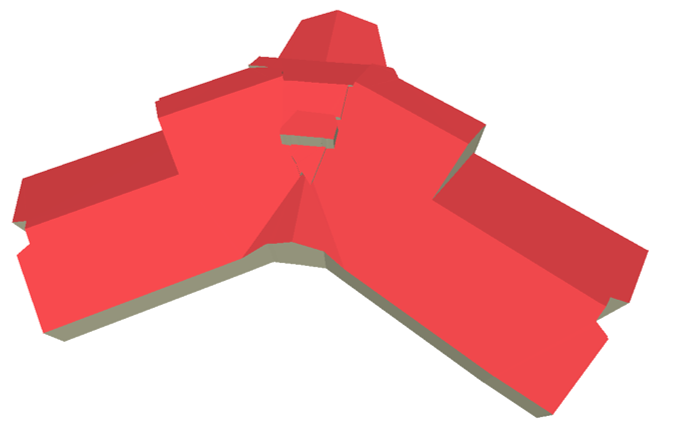}
		\label{fig:lod2_cmp_1e}
	\end{subfigure}
	
	\begin{subfigure}[t]{0.18\textwidth}
		\includegraphics[width=\linewidth]{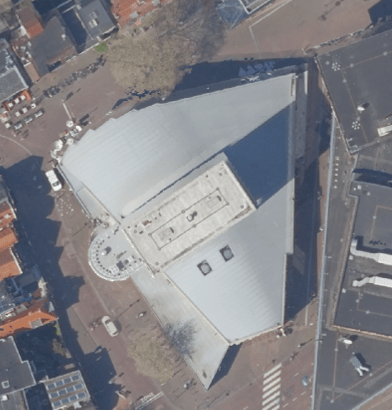}
		\label{fig:lod2_cmp_2a}
	\end{subfigure}
	\hspace*{\fill}
	\begin{subfigure}[t]{0.18\textwidth}
		\includegraphics[width=\linewidth]{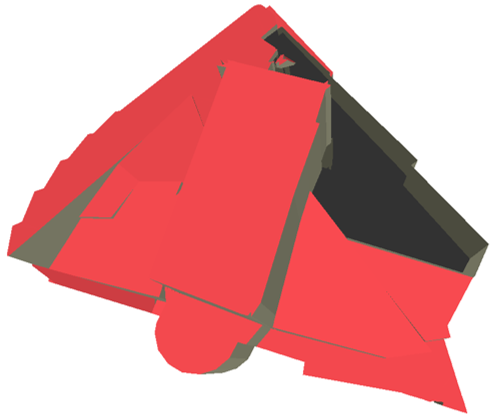}
		\label{fig:lod2_cmp_2b}
	\end{subfigure}
	\hspace*{\fill}
	\begin{subfigure}[t]{0.18\textwidth}
		\includegraphics[width=\linewidth]{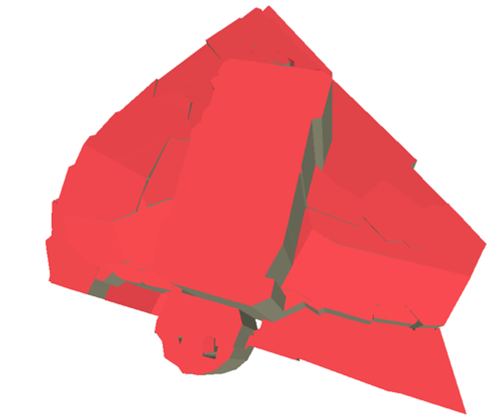}
		\label{fig:lod2_cmp_2c}
	\end{subfigure}
	\hspace*{\fill}
	\begin{subfigure}[t]{0.18\textwidth}
		\includegraphics[width=\linewidth]{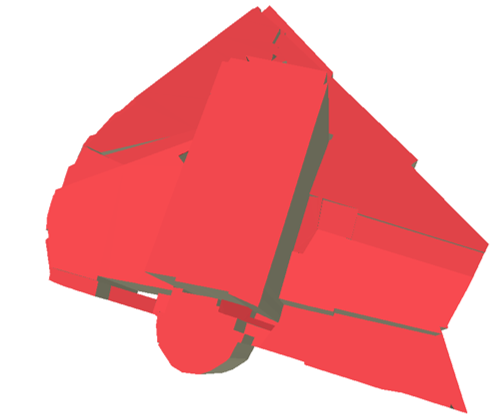}
		\label{fig:lod2_cmp_2d}
	\end{subfigure}
	\hspace*{\fill}
	\begin{subfigure}[t]{0.18\textwidth}
		\includegraphics[width=\linewidth]{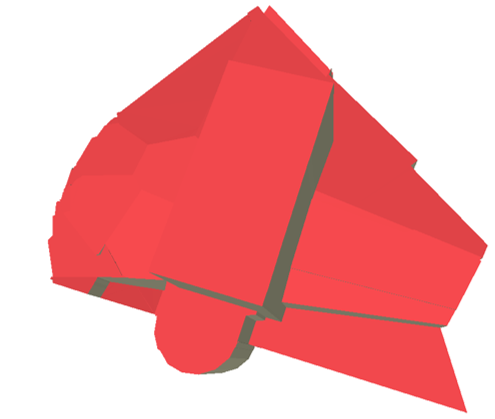}
		\label{fig:lod2_cmp_2e}
	\end{subfigure}
	
	\begin{subfigure}[t]{0.18\textwidth}
		\includegraphics[width=\linewidth]{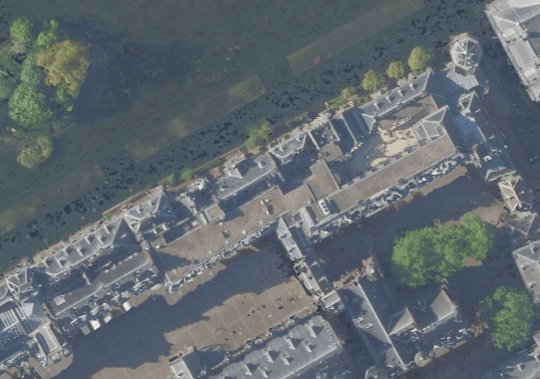}
		\label{fig:lod2_cmp_3a}
	\end{subfigure}
	\hspace*{\fill}
	\begin{subfigure}[t]{0.18\textwidth}
		\includegraphics[width=\linewidth]{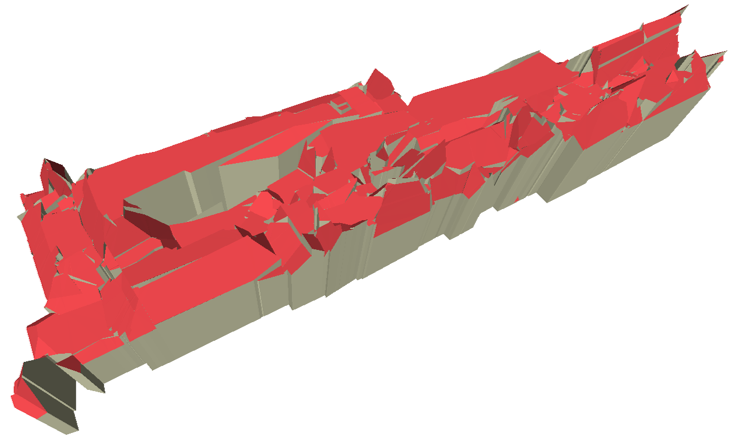}
		\label{fig:lod2_cmp_3b}
	\end{subfigure}
	\hspace*{\fill}
	\begin{subfigure}[t]{0.18\textwidth}
		\includegraphics[width=\linewidth]{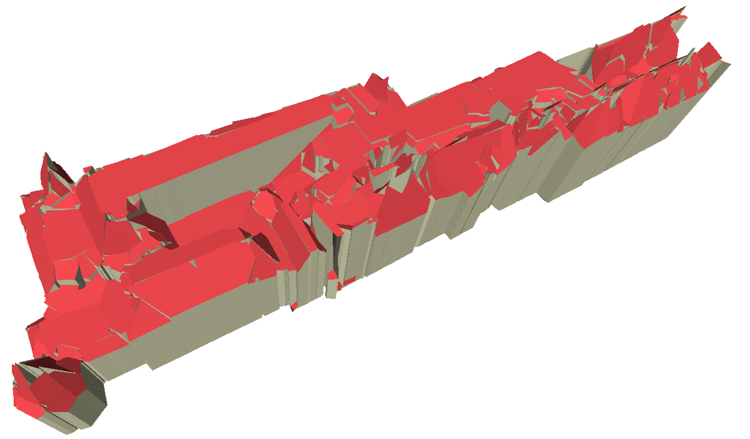}
		\label{fig:lod2_cmp_3c}
	\end{subfigure}
	\hspace*{\fill}
	\begin{subfigure}[t]{0.18\textwidth}
		\includegraphics[width=\linewidth]{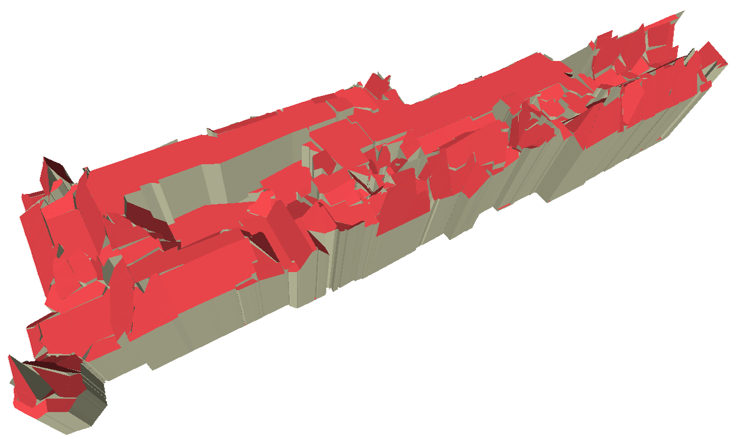}
		\label{fig:lod2_cmp_3d}
	\end{subfigure}
	\hspace*{\fill}
	\begin{subfigure}[t]{0.18\textwidth}
		\includegraphics[width=\linewidth]{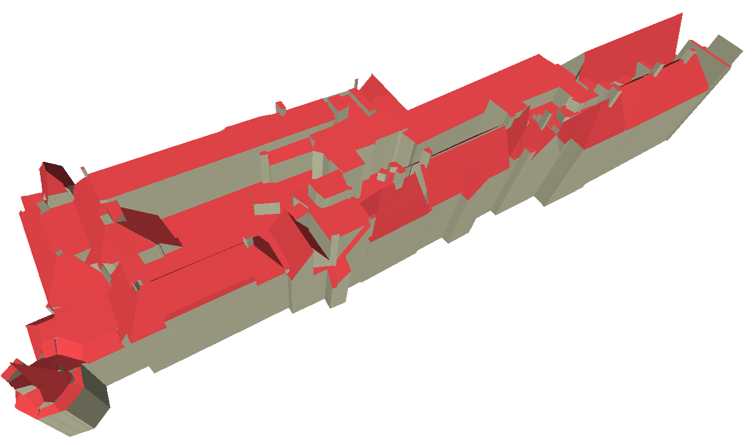}
		\label{fig:lod2_cmp_3e}
	\end{subfigure}
	
	\begin{subfigure}[t]{0.18\textwidth}
		\includegraphics[width=\linewidth]{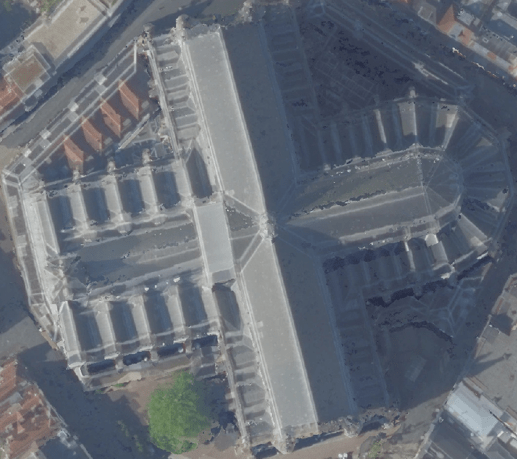}
		\caption{}
		\label{fig:lod2_cmp_4a}
	\end{subfigure}
	\hspace*{\fill}
	\begin{subfigure}[t]{0.18\textwidth}
		\includegraphics[width=\linewidth]{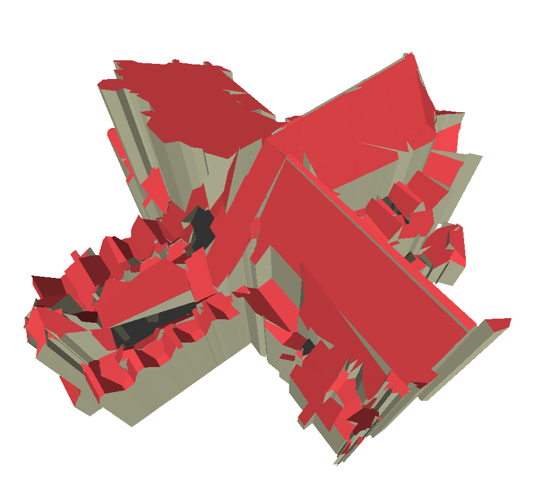}
		\caption{}
		\label{fig:lod2_cmp_4b}
	\end{subfigure}
	\hspace*{\fill}
	\begin{subfigure}[t]{0.18\textwidth}
		\includegraphics[width=\linewidth]{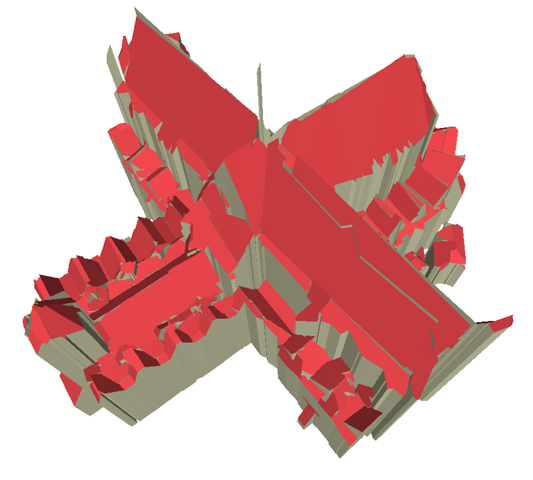}
		\caption{}
		\label{fig:lod2_cmp_4c}
	\end{subfigure}
	\hspace*{\fill}
	\begin{subfigure}[t]{0.18\textwidth}
		\includegraphics[width=\linewidth]{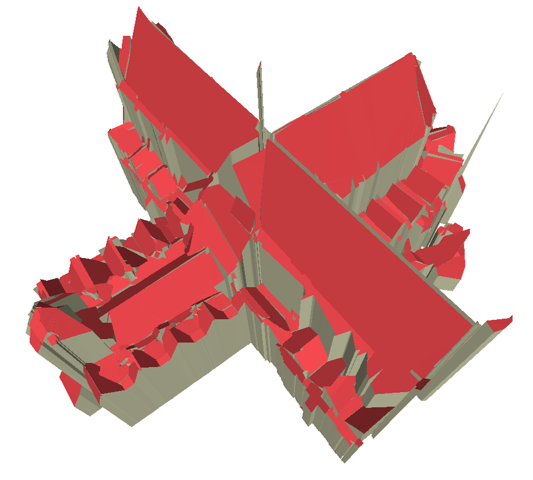}
		\caption{}
		\label{fig:lod2_cmp_4d}
	\end{subfigure}
	\hspace*{\fill}
	\begin{subfigure}[t]{0.18\textwidth}
		\includegraphics[width=\linewidth]{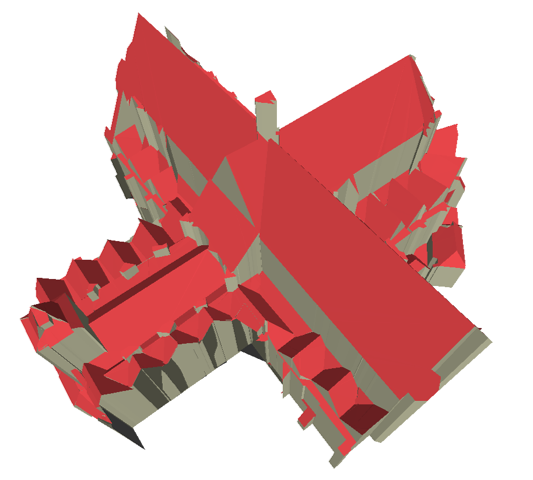}
		\caption{}
		\label{fig:lod2_cmp_4e}
	\end{subfigure}

	\caption{Comparison of LoD2 Building Models: Input building images are shown in (a), while the reconstruction results using HEAT generated rooflines are shown in (b), plane detection method in (c), and our line segments in (d). Ground truth 3D BAG models reconstructed from LiDAR point clouds and plane detection are shown in (e).} 
	\label{fig:lod2_cmp}
\end{figure}

\paragraph{\textbf{Evaluation of reconstructed buildings.}}\label{buidling_eval}
We selected twelve buildings from the three cities that were previously identified and used them to reconstruct LoD2 models and evaluate our results.
We used 3D BAG building models~\cite{peters2022automated} reconstructed from LiDAR point clouds as the ground truth for evaluation. To compare the reconstructed building models, we utilized the same image-dense matching generated point clouds as input and applied three different approaches for roofline generation: 1) plane detection from the dense matching point cloud, 2) HEAT, and 3) our method (i.e., KIPPI).
The evaluation criteria focused on model complexity, Hausdorff distance~\cite{aspert2002mesh}, and RMSE. We measured the complexity of the models based on the number of reconstructed faces and calculated the Hausdorff distance using sampled points from the building models. We computed the average Hausdorff distance and root mean square error (RMSE) of all the selected buildings in comparison to their respective ground truth models. It should be noted that if a model contains more faces with larger errors, then it is likely to have redundant faces. Conversely, models with smaller errors and more faces tend to capture more details.

\begin{table}[H]
	\centering
	\begin{tabularx}{0.6\textwidth}{>{\centering\arraybackslash}X>{\centering\arraybackslash}X>{\centering\arraybackslash}X>{\centering\arraybackslash}X}
		\toprule
		& \textbf{Faces} & \textbf{Mean} & \textbf{RMSE} \\
		\midrule
		\textbf{3D BAG} & 1762.5 & - & - \\
		\midrule
		\textbf{HEAT} & 3150.75 & 0.67 & 1.16 \\
		\midrule
		\textbf{Planes} & 5660.08 & 0.49 & 0.67 \\
		\midrule
		\textbf{KIPPI} & 4356.33 & \textbf{0.38} & \textbf{0.65} \\
		\bottomrule
	\end{tabularx}%
	\caption{A comprehensive quantitative evaluation of LoD2 building models using the averaged number of faces, mean Hausdorff distance, and RMSE metrics.}
	\label{tab:tab_com}%
\end{table}%

Table~\ref{tab:tab_com} and Figure~\ref{fig:lod2_cmp} show the quantitative and qualitative results of the different methods, respectively. Our method produced results with higher accuracy and completeness than the HEAT method~\cite{chen2022heat}, which had the lowest complexity but the highest error due to missing line segment completeness. Furthermore, the reconstructed building models using our rooflines exhibited lower complexity and lower errors compared to the rooflines obtained from the plane detection method~\cite{peters2022automated}. Figure~\ref{fig:lod2_cmp_4d} demonstrate that even when orthophoto images are somewhat blurry, our method can maintain high accuracy and low complexity, indicating its high robustness.
It is important to mention that in KIPPI~\cite{bauchet2018kippi}, the gradient magnitude parameter can be adjusted to effectively handle low-intensity values in shadowed regions. Moreover, in the case of textureless regions lacking clear edge information, such as the example depicted in Figure~\ref{fig:roofline_cmp}c, the curved surface nature of the roof can result in the loss of certain rooflines during the detection phase.

\begin{figure}[!tb]
	\centering
	\begin{subfigure}[t]{0.48\textwidth}
		\includegraphics[width=\linewidth]{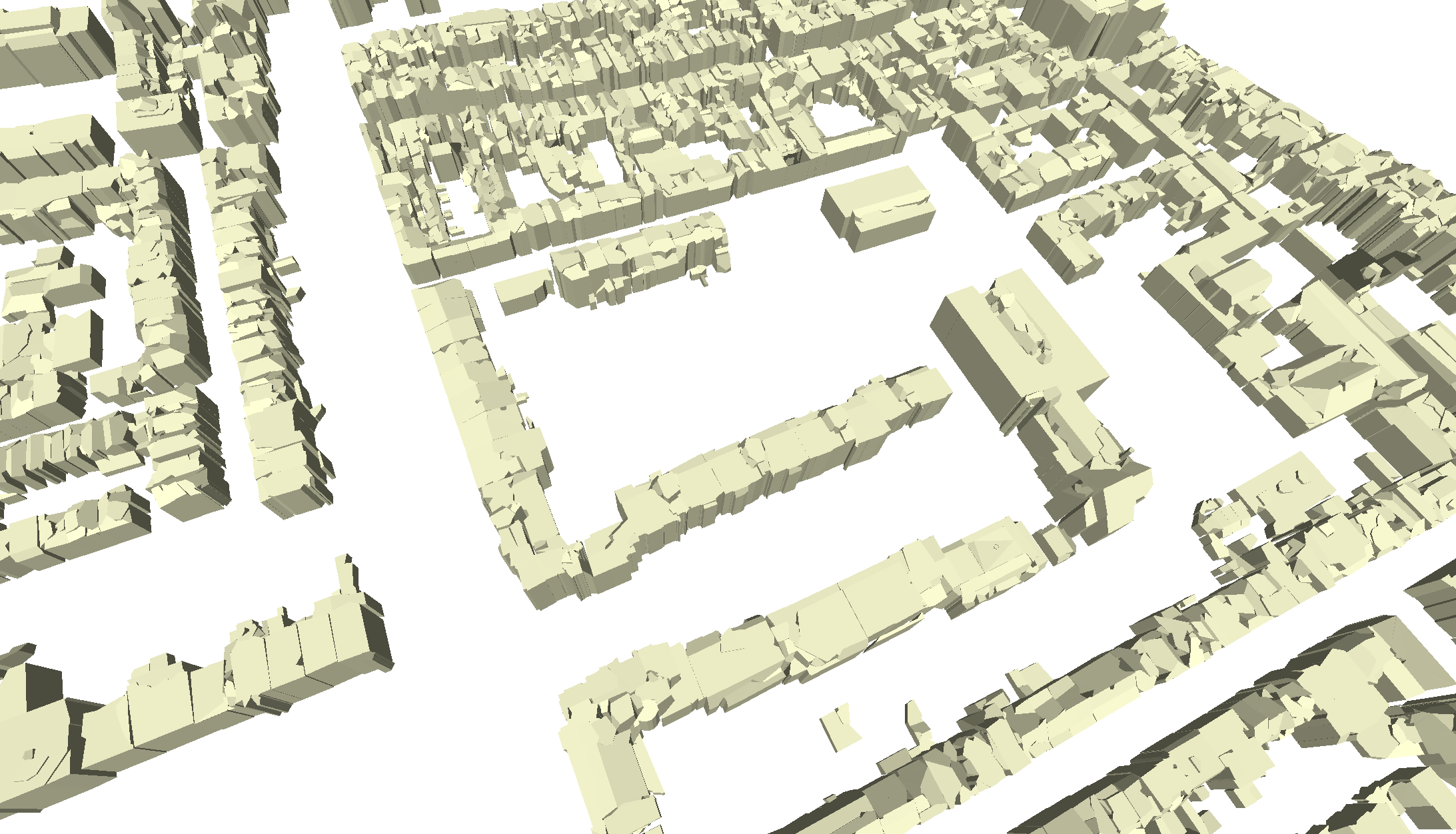}
		\caption{}
		\label{fig:lg_lod2_a}
	\end{subfigure}
	\hspace*{\fill}
	\begin{subfigure}[t]{0.48\textwidth}
		\includegraphics[width=\linewidth]{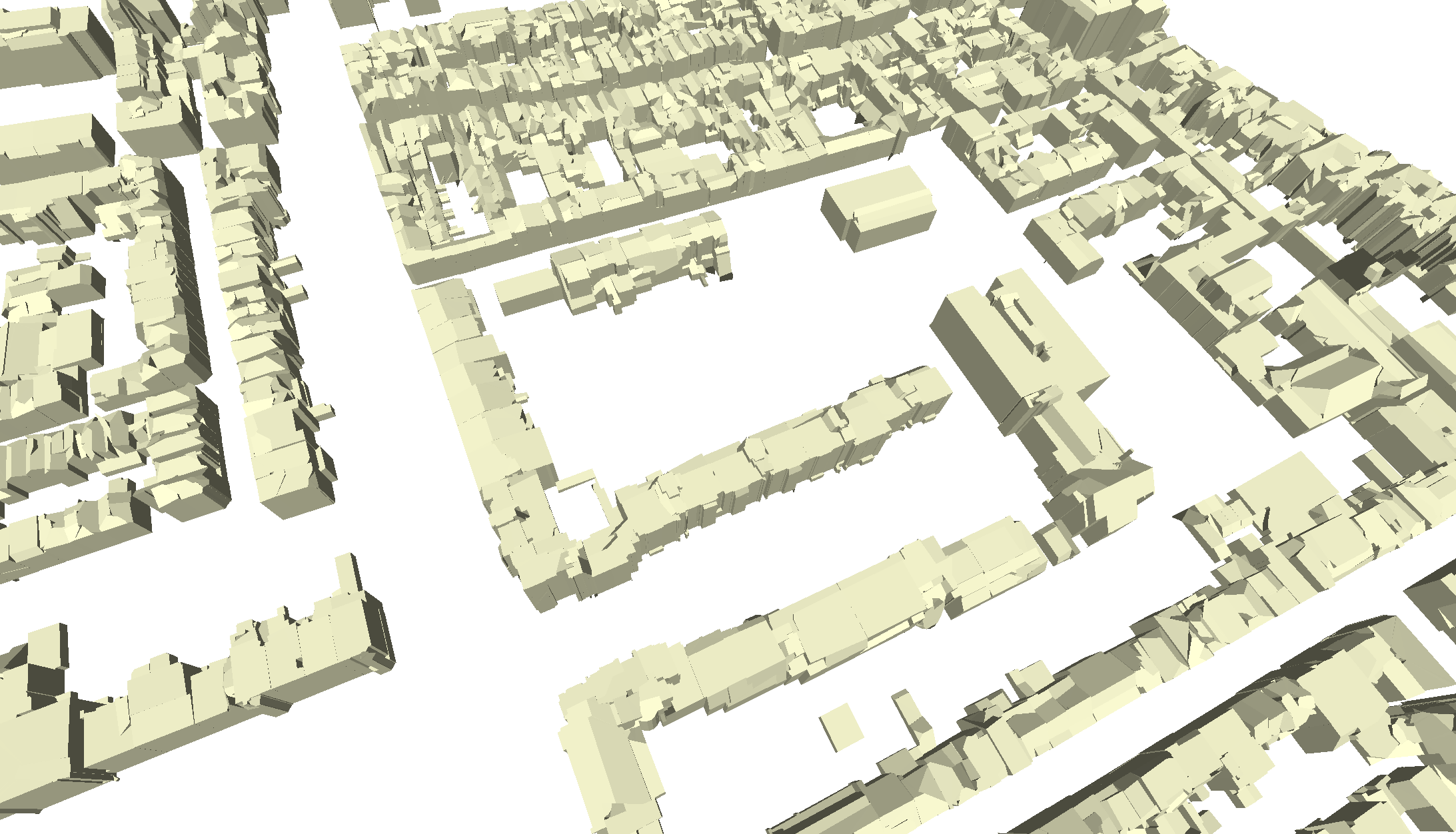}
		\caption{}
		\label{fig:lg_lod2_b}
	\end{subfigure}
	\caption{Comparison of large-scale LoD2 building models reconstructed from extracted rooflines and image dense matching generated point clouds. 
		(a) Building models reconstructed using rooflines extracted from plane detection.
		(b) Building models reconstructed using rooflines extracted from our method(i.e., KIPPI method).
        } 
	\label{fig:lg_lod2}
\end{figure}

Figure~\ref{fig:lg_lod2} illustrates the outcomes of the LoD2 building model reconstruction for a large-scale urban scene. Notably, our method demonstrates its capacity to be seamlessly applied to such scenes, as depicted in Figure~\ref{fig:lg_lod2_b}. This suggests its potential as a viable alternative to image-based approaches for reconstructing the complete 3D BAG building model. Additionally, we observe that our method produces building models with enhanced geometric regularity compared to the plane detection-based method (see Figure~\ref{fig:lg_lod2_b}).

\section{Conclusion}
In conclusion, this paper presented a method for extracting rooflines using line detection, which is used for the improved reconstruction of building models at the LoD2 level from point clouds generated from images. The approach has the advantage of being able to extract relatively complete rooflines without the need for any pre-labeled training data or pre-trained models. Furthermore, the method surpassed both traditional plane detection-based methods and state-of-the-art deep learning methods based on transformers in terms of both the accuracy and completeness of the reconstructed building. The method is applicable to large-scale urban scenes and can be used for extracting roof structures of complex buildings. Although deep learning methods have received increasing attention for extracting roof structures from images, they still require a large amount of manually labeled training data and cannot ensure the geometric completeness of the extracted roof structures, especially for buildings with complex structures. The proposed method has the potential to overcome these limitations and improve the LoD2 model reconstruction from dense image matching point clouds which can then become a promising alternative for automated LoD2 building model reconstruction from LiDAR. Further research could be directed toward developing more accurate algorithms for line detection from low-resolution images, such as satellite imagery, to improve the reconstruction of building models also from those sources.

\section{Acknowledgement}
This project has received funding from the European Research Council (ERC) under the Horizon Europe Research \& Innovation Programme (grant agreement no. 101068452 3DBAG: detailed 3D Building models Automatically Generated for very large areas).

%
%

\end{document}